# Hallucinations and Truth:A Comprehensive Accuracy Evaluation of RAG,LoRA and DoRA

Mohammad Baqar (baqar22@gmail.com) & Rajat Khanda (rajat.mnnit@gmail.com)

## Abstract

Recent advancements in Generative AI have significantly improved the efficiency and adaptability of natural language processing (NLP) systems, particularly through Retrieval-Augmented Generation (RAG), Low-Rank Adaptation (LoRA), and Weight-Decomposed Low-Rank Adaptation (DoRA). RAG integrates external knowledge to enhance factual consistency in generative outputs, while LoRA enables parameter-efficient fine-tuning of large language models (LLMs). DoRA further refines this process by optimizing fine-tuning through adaptive parameter ranking and domain-aware weight adjustments, improving learning efficiency while maintaining inference performance. This paper presents a large-scale empirical evaluation of RAG, LoRA, and DoRA, with model fine-tuning and generation performance assessed on 20,000 FAQ-based queries, while the knowledge base spans 400,000 entries. The study analyzes key performance metrics such as accuracy, relevance, and inference latency. Experimental results demonstrate that DoRA achieves the highest accuracy (90.1%), relevance score (0.88), and lowest latency (110 ms per query), outperforming both LoRA and RAG in real-world, domain-specific generative AI applications. Furthermore, this study examines the trade-offs between fine-tuning efficiency, computational cost, and real-time adaptability across different models. Findings highlight RAG's effectiveness in knowledge grounding, LoRA's cost-efficient domain adaptation, and DoRA's ability to balance fine-tuning efficiency with model precision. These insights provide practical guidance for deploying AI-driven generative systems in accuracy-critical domains such as healthcare, finance, and legal services, ensuring scalability, reliability, and optimal performance in dynamic environments.

## 1. Introduction

**Retrieval-Augmented Generation (RAG), Low-Rank Adaptation (LoRA), and Weight-Decomposed Low-Rank Adaptation (DoRA)** represent transformative advancements in **AI-driven retrieval and generation**, addressing critical challenges in **natural language processing (NLP)**. RAG enhances generative models by dynamically retrieving external knowledge, improving factual accuracy for applications in healthcare, legal services, and customer support. LoRA introduces a scalable, cost-efficient fine-tuning approach, enabling domain-specific adaptation of **large language models (LLMs) with minimal computational overhead**. **DoRA** extends **LoRA** by optimizing model fine-tuning weight decomposition, and efficient gradient updates, enhancing task-specific learning while minimizing computational overhead.significantly improving contextual consistency and response precision in **high-stakes domains such as regulatory compliance and scientific research**.

Despite their strengths, these systems face challenges such as hallucinations (factually incorrect outputs), retrieval misalignment, and cost-performance trade-offs. RAG's reliance on external knowledge can introduce errors if retrieval is flawed, while LoRA and DoRA's effectiveness depends on parameter selection, dataset quality, and fine-tuning optimization. To address these challenges, this study conducts a large-scale evaluation of RAG, LoRA, and DoRA, assessing generative performance on 20,000 FAQ-based queries, while evaluating knowledge retrieval effectiveness across 400,000 entries. The benchmarking covers key metrics such as accuracy, precision, recall, BLEU score, relevance, and latency. Results demonstrate that DoRA outperforms RAG and LoRA, achieving the highest accuracy (90.1%) and relevance score (0.88), while reducing inference latency to 110 ms per query. These findings provide valuable insights into optimizing generative AI systems for accuracy-critical applications in dynamic, high-stakes environments.

Furthermore, this paper explores strategies for enhancing **retrieval efficiency in RAG, hallucination mitigation, and cost-performance trade-offs**, including **hybrid retrieval and re-ranking techniques and parameter-efficient tuning,**. DoRA bridges the gap between **parameter-efficient fine-tuning (PEFT) and full fine-tuning (FT)** while ensuring scalability for real-world AI applications. This research provides practical insights into **deploying RAG, LoRA, and DoRA systems in accuracy-critical domains**, ensuring their **reliability, efficiency, and adaptability** in dynamic environments.

## 2. Advancements in RAG, LoRA, and DoRA Systems: Challenges and Innovations

The rapid evolution of RAG, LoRA, and DoRA systems has significantly advanced natural language processing (NLP), addressing limitations of traditional large language models (LLMs). RAG integrates dynamic retrieval mechanisms to fetch external knowledge, enabling factually accurate and contextually grounded outputs, excelling in domains such as legal advisory, medical diagnostics, and customer support. However, challenges like retrieval quality and alignment persist, where irrelevant or noisy content can lead to hallucinations or incoherent responses. Innovations like hybrid retrieval techniques—combining sparse (e.g., BM25) and dense (e.g., DPR) approaches—have improved retrieval precision and relevance[3][4]. LoRA optimizes fine-tuning for LLMs by selectively adapting a small subset of parameters, reducing computational costs while maintaining performance, making it ideal for domain-specific applications. However, challenges remain in balancing parameter selection, dataset quality, and model performance. Techniques such as reinforcement learning from human feedback (RLHF) and structured parameter tuning have enhanced LoRA's adaptability and reliability[6][7]. DoRA enhances LoRA by refining model fine-tuning through weight decomposition, optimizing domain adaptation while preserving computational efficiency. DoRA introduces structured decomposition techniques to allocate adaptation parameters better, reducing redundancy and improving generalization across domain-specific tasks. Recent advancements in adaptive rank decomposition has mitigated some issues, with studies showing significant improvements in fine-tuning efficiency and domain alignment through structured low-rank adaptations and specialized embeddings[8][11]. These advancements highlight the potential of AI-driven fine-tuning strategies, but continued research is needed to optimize hybrid tuning methods, rank-efficient adaptation strategies, and domain-specific customization for robust performance across applications.

### 2.1 Foundations and Progress in RAG Accuracy, LoRA Development, and DoRA Expansion

Research into RAG, LoRA, and DoRA systems has highlighted their potential to address hallucinations, factual inconsistencies, and inefficiencies in domain-specific adaptation.

**RAG Systems**: Hallucination remains a critical issue in generative AI, where models produce fabricated outputs by over-relying on training data patterns[1]. RAG mitigates this by integrating external retrieval components that fetch relevant information from trusted knowledge bases, grounding outputs in factual evidence[2]. Dense retrieval methods, such as Dual Encoder models and Approximate Nearest Neighbor (ANN) search, have enhanced retrieval precision and relevance[9][10]. However, retrieval quality remains crucial, as irrelevant documents can disrupt generative processes, necessitating advanced filtering and re-ranking techniques[11].

**LoRA**: LoRA achieves domain-specific adaptation with reduced computational costs by updating only a subset of model parameters. Research shows that LoRA fine-tuning delivers comparable performance to full fine-tuning for tasks like legal or medical applications, making it accessible for resource-constrained organizations[12][13].

**DoRA**: DoRA enhances LoRA techniques by introducing weight decomposition, allowing for more efficient fine-tuning of LLMs in domain-specific applications. It reduces parameter redundancy and improves adaptation efficiency, making it highly effective for tasks requiring precision tuning, such as legal and financial document analysis. Compared to generic LoRA implementations, DoRA achieves up to 25% reduction in adaptation errors while maintaining computational efficiency [14]. However, challenges persist, including scalability across diverse datasets, potential overfitting to niche domains, and ensuring adaptation stability across different model architectures [15]. Standard benchmarks like GLUE, BEIR, and TREC evaluate fine-tuning effectiveness, while domain-specific datasets like LegalQA and FinBERT datasets assess DoRA's precision in specialized fields [16][17]. Ongoing research aims to enhance hybrid tuning techniques, improve multimodal adaptation, and refine cost-performance trade-offs to maximize DoRA's real-world impact. These advancements highlight the need for further optimization in structured low-rank adaptation methods to ensure robustness and efficiency in practical applications.

### 2.2 Enhancing Retrieval and Generative Alignment in RAG Systems

The effectiveness of Retrieval-Augmented Generation (RAG) systems depends on two critical components: retrieval quality and generative alignment. Optimizing both is essential for achieving high accuracy and reliability, especially in domain-specific tasks requiring factual correctness.

**Enhancing Retrieval Quality**:

Retrieval quality is measured using precision (P), recall (R), and F1-score ($F_1$), defined as:

$$P = \text{True Positives}/(\text{True Positives}+\text{False Positives})$$

$$R = \text{True Positives}/(\text{True Positives}+\text{False Negatives})$$

$$F1 = 2\times(P\times R)/(P+R)$$

Dense retrievers like DPR and ColBERT use embeddings to encode queries and documents in a shared vector space, enabling efficient retrieval via Approximate Nearest Neighbor (ANN) search. These methods improve retrieval relevance by 25-30% compared to sparse retrievers like BM25[3][4]. Hybrid approaches combining sparse and dense retrieval further enhance performance, reducing false negatives and improving F1-scores by up to 20%.

**Generative Alignment with Retrieved Information**:

Alignment challenges arise when models neglect key retrieved information or fabricate content.

Attention-based fusion mechanisms and re-ranking strategies address these issues. For example, re-rankers trained on Natural Questions datasets reduce irrelevant retrievals by 15%, improving response accuracy by 12%.

**Case Study and Results**: A RAG system combining FAISS search and Llama 3.1 prompt based reranking achieved an F1-score of 92% on TriviaQA, surpassing standalone LLMs by 15%. Retrieval precision and recall reached 89% and 93%, respectively, demonstrating the system's ability to integrate retrieval and generation effectively.

### 2.3 LoRA: Efficient Adaptation for Domain-Specific Applications

LoRA offers a computationally efficient method for fine-tuning large-scale neural networks, particularly for domain-specific tasks. Traditional fine-tuning requires updating all parameters, which is resource-intensive. LoRA addresses this by introducing low-rank matrices to approximate updates, significantly reducing trainable parameters.

**Core Concept of LoRA**: LoRA decomposes the weight update matrix $\Delta W$ into two smaller matrices:

$$\Delta W = A \times B$$

Here, $A$ (dimensions $d \times r$) and $B$ (dimensions $r \times d$) are low-rank matrices, where $r \ll d$. This reduces the number of parameters from $d^2$ to $r \times (d+d)$, making LoRA highly efficient. For example, with $r=8$ and $d=768$, $d^2 = 768^2 = 589{,}824$ parameters.

With LoRA and r=8:

**rx(d+d)** = 8×(768+768)= 12,288 parameters

**Applications in Domain-Specific Adaptation**: LoRA is particularly effective for adapting pre-trained models to specialized domains such as legal, medical, or technical fields. Fine-tuning with LoRA involves training only the low-rank matrices A and B while keeping the original weights frozen. Additionally, it modifies the forward pass during inference to integrate the updates from the low-rank matrices, enabling domain-specific customization without the need to retrain the entire model.

$$W' = W + (A \times B)$$

Here, $W$ is the frozen pre-trained weight matrix, and $W'$ incorporates domain-specific knowledge.

**Performance Metrics**: The performance metrics show a 20% improvement in F1-score for domain-specific accuracy on legal corpora. The memory footprint is reduced by 75% compared to full fine-tuning. Additionally, training time is accelerated by 3 to 5 times due to fewer trainable parameters.

A **practical example** of fine-tuning an LLM for medical question answering (QA) using LoRA shows that only 0.1% to 1% of the model's parameters are updated. This approach leads to a 15% improvement in accuracy on PubMed datasets compared to zero-shot performance.

**Hybrid Use with RAG**: Integrating LoRA with RAG optimizes domain adaptation by combining efficient fine-tuning with enhanced factual retrieval. While RAG improves accuracy by retrieving relevant external knowledge, LoRA ensures the generative model aligns with domain-specific nuances, addressing both retrieval and text generation challenges. LoRA's low-rank approach preserves the base model's capabilities while minimizing resource demands, making it a cost-effective solution for specialized applications. When used together, LoRA enhances the accuracy, adaptability, and real-world applicability of AI systems, ensuring both efficiency and contextual relevance in complex domains

### 2.4 Weight-Decomposed Low-Rank Adaptation (DoRA): Efficient Domain-Specific Fine-Tuning

Weight-Decomposed LowRank Adaptation (DoRA) addresses the accuracy gap between LoRA and full fine-tuning while maintaining computational efficiency. By decomposing pre-trained weights into magnitude and direction components and employing LoRA for directional updates, DoRA enhances learning capacity without additional inference overhead. The method demonstrates superior performance in fine-tuning LLaMA, LLaVA, and VL-BART across various tasks including commonsense reasoning and visual instruction tuning, representing a significant advancement in parameter-efficient model adaptation[18].

### 2.5 Benchmarks, Evaluation Metrics, and Addressing Cost-Performance Tradeoffs

LoRA, and DoRA systems optimize large language models by integrating weight decomposition and low-rank adaptation techniques, enabling parameter-efficient fine-tuning without compromising performance. RAG systems are assessed based on retrieval quality and generative accuracy, using metrics like Precision, Recall, and F1-score to measure the relevance and completeness of retrieved information. Generative accuracy is evaluated through BLEU, ROUGE, and METEOR scores, which assess coherence and alignment with reference outputs. Benchmarks such as Natural Questions (NQ) and TriviaQA gauge factual accuracy, while BEIR and TREC measure retrieval performance[3][4][5]. LoRA systems prioritize efficiency, with key metrics including domain-specific accuracy, F1-score, perplexity, and parameter efficiency, achieving a 75% reduction in computational overhead while maintaining near state-of-the-art accuracy[6][7]. DoRA systems emphasize low-rank adaptation and weight decomposition rather than direct retrieval improvements. Evaluations focus on Domain Adaptation Coherence (DAC) and Relevance Fusion Scores to measure alignment between adapted model parameters and domain-specific knowledge. Efficiency is quantified through FLOP reduction and compression rates, ensuring DoRA maintains high generation quality while significantly reducing fine-tuning costs. By decomposing weight matrices and selectively adapting rank components, DoRA achieves improved adaptation precision without requiring full model retraining.

Incremental updates to knowledge bases minimize re-training needs in RAG, enhancing adaptability in dynamic environments. LoRA systems leverage selective updates to low-rank matrices, reducing memory usage and computational requirements by up to 75%, with fine-tuning requiring less than 1% of the original parameters, enabling high accuracy in specialized domains at a lower resource cost[6][7]. In healthcare, RAG, LoRA, and DoRA systems improve accuracy, compliance, and rapid response times for clinical decision support. In legal services, these systems enhance retrieval and alignment for contract analysis, reducing errors and improving efficiency. In education, they provide real-time support for personalized learning, adapting dynamically to student interactions. By leveraging scalable architectures, efficient adaptation methods, and retrieval-aligned fine-tuning, these AI-driven approaches deliver robust, cost-effective solutions for accuracy-critical applications, ensuring reliable performance across various high-stakes environments.

## 3. Methodology

This study evaluates the accuracy, reliability, and applicability of LoRA, RAG, and DoRA systems using a comprehensive framework. The methodology incorporates diverse metrics and benchmarks tailored to generative and retrieval tasks.

### 3.1 Experimental Setup and System Configuration

The experimental setup evaluates **LoRA, RAG, and DoRA** systems to ensure a robust analysis across retrieval and generation tasks. **LoRA** provides an efficient fine-tuning approach for large pre-trained models, optimizing adaptation while maintaining computational efficiency. **RAG** integrates dense retrieval techniques, such as Dual Encoders and Approximate Nearest Neighbor (ANN) search, with **LoRA** fine-tuned generative models, assessing the interaction between retrieval quality and generation accuracy.

To benchmark retrieval and ranking performance of RAG we conducted an extensive evaluation using 400,000 technical troubleshooting FAQs as knowledge base across different retrieval methods. Retrieval and ranking capabilities were assessed using 20,000 technical service tickets. The study measured key performance indicators, including precision, accuracy, latency, and relevancy, to evaluate system efficiency under real-world conditions. Results demonstrated that DoRA outperforms both LoRA and RAG across critical metrics, achieving the highest accuracy (90.1%), a relevance score of 0.88, and the lowest latency (110 ms per query). These findings highlight DoRA's effectiveness in optimizing model fine-tuning while maintaining computational efficiency, making it particularly well-suited for high-stakes applications in healthcare, finance, and legal services.

- **Higher accuracy (90.1%)** compared to **LoRA (85.5%)** and **RAG (81.2%)**
- **Improved relevance scores (0.88) over LoRA (0.85) and RAG (0.84)**
- **Lower latency (110ms) than RAG (150ms) while maintaining high knowledge coverage (98%)**

This dataset-driven evaluation provides critical insights into how retrieval optimizations impact RAG's generative performance, and the parameter efficient finetuning reduces hallucination rates and improves contextual reliability. The following sections will further analyze these

#### 3.1.1 Datasets

To evaluate the performance of **Low-Rank Adaptation (LoRA)**, **Retrieval-Augmented Generation (RAG)**, and **Decomposed Low-Rank Adaptation (DoRA)** systems, a large-scale dataset of technical troubleshooting FAQs was utilized. The dataset consisted of **400,000** frequently asked questions (FAQs) covering a diverse range of troubleshooting scenarios. These FAQs served as the primary knowledge base for the RAG system and were also used to fine-tune both the LoRA and DoRA models, ensuring a consistent basis for comparison.

For evaluation, a separate test set of **20,000** questions was employed. This dataset was specifically designed to assess the models' ability to generate accurate and contextually relevant responses. The evaluation process measured key performance metrics, including retrieval accuracy, response relevance, and overall system effectiveness in real-world troubleshooting scenarios.

### 3.2 Evaluation Metrics for Generative and Retrieval Systems

A comprehensive suite of evaluation metrics was employed to assess the performance of both retrieval-based and generative systems, focusing on three primary aspects: **quality**, **relevance**, and **efficiency**.

For **retrieval-based systems**, the evaluation emphasized ranking effectiveness and retrieval precision. The following metrics were utilized:

- **Mean Reciprocal Rank (MRR)** – Assesses the ability of the system to return the most relevant document at the highest rank.
- **Normalized Discounted Cumulative Gain (NDCG)** – Measures the ranking quality by considering the relevance of retrieved documents and their positions in the result set.
- **Retrieval Precision** – Evaluates the proportion of correctly retrieved relevant documents compared to the total retrieved set.

For **generative systems**, the assessment focused on the accuracy and relevancy of the generated responses. The following metrics were employed:

- **Accuracy** – Determines the correctness of the generated response compared to the expected answer.
- **Relevance Score** – Captures how well the generated response aligns with the user query and intent.
- **Bilingual Evaluation Understudy (BLEU)** – Measures textual overlap between the generated response and reference answers, commonly used for evaluating text generation quality.

By leveraging these evaluation metrics, the performance of LoRA, DoRA, and RAG systems was systematically analyzed to determine their effectiveness in both retrieval and generative tasks.

### 3.3 Standalone LoRA, RAG, and DoRA: Use Cases and Performance Insights

**Standalone LoRA** is highly efficient for tasks that require quick adaptation while operating under resource constraints. However, its lack of access to dynamic or external knowledge limits its applicability in real-time scenarios, making it less suitable for applications requiring continuous updates or external context integration.
**RAG with Static Retrieval** integrates retrieval mechanisms to enhance the knowledge base of standalone models, mitigating their static limitations. However, it struggles with ambiguous queries or noisy datasets, as retrieval quality depends heavily on the relevance and accuracy of pre-indexed content, making real-time adaptability challenging.
**DoRA's** key contributions include adaptive relevance refinement, where it optimizes query interpretation, making it highly effective for complex, multi-turn interactions. Its efficient architecture leverages weight decomposition and incremental fine-tuning, reducing computational overhead while preserving high precision.

### 3.4 Hallucination Analysis

**LoRA is prone** to hallucinations due to static pre-trained knowledge. Hallucinations may arise from noisy or irrelevant document retrieval in **RAG**.
**DoRA** minimizes hallucination rates by leveraging low-rank adaptation techniques to enhance context comprehension and fine-tuning efficiency. Its architecture integrates structured knowledge constraints and adaptive parameter tuning, ensuring that generated outputs remain aligned with the underlying data distribution while mitigating misinformation.

### 3.5. Performance Benchmarking

To evaluate accuracy, relevancy, and latency, we conducted a study assessing the performance of LoRA, RAG, and DoRA using a dataset of 20,000 technical service tickets. LoRA and DoRA were fine-tuned on a corpus of 400,000 technical troubleshooting FAQs, while RAG utilized the same set of FAQs as its knowledge base.The results, summarized in Table 1, demonstrate that DoRA outperforms LoRA and RAG across multiple dimensions, particularly in areas such as accuracy, efficiency, and scalability. This study highlights the significant impact of incorporating domain-specific data in improving the generation quality of **RAG** models while demonstrating **LoRA's** efficiency in fine-tuning and **DoRA's** structured parameter adaptation for complex knowledge-intensive tasks.

**Table1: Performance Comparison RAG, LoRa and Dora**

| System | Accuracy (%) | Relevance Score | Latency (ms) | Coverage (%) |
|---|---|---|---|---|
| **Standard RAG** | 81.2 | 0.84 | 150 | 90 |
| **LoRA** | 85.5 | 0.85 | 120 | 95 |
| **DoRA** | **90.1** | **0.88** | **110** | **98** |

**Observations**:

**Accuracy Gains:** DoRA achieves 90.1% accuracy, marking an 8.9% improvement over RAG and a 4.6% gain over LoRA. The enhanced accuracy is attributed to DoRA's structured parameter adaptation and fine-grained control over model updates.

**Relevance Score:** DoRA achieves a 0.88 relevance score, ensuring domain-adaptive fine-tuning aligns more effectively with user intent.

**Latency Reduction:** DoRA achieves a 110 ms response time, 26.7% faster than RAG (150 ms) and 8.3% faster than LoRA (120 ms), due to its efficient weight decomposition and optimized inference pathways.

**Coverage Expansion**: DoRA outperforms both RAG and LoRA in coverage, retrieving relevant information in **98% of cases**. This **8% increase over RAG** and **3% increase over LoRA** ensures that a broader set of queries receive accurate and well-ranked responses.

DoRA significantly enhances efficiency compared to both RAG and LoRA, demonstrating higher accuracy, better relevance alignment, lower latency, and broader adaptability. These improvements make DoRA particularly suited for real-time, high-precision applications such as legal analysis, medical decision support, and enterprise knowledge management. Future research should explore further optimizations in adaptive fine-tuning and model efficiency to enhance DoRA's scalability and responsiveness in dynamic environments.

### 3.5.1 Generative Output Quality

For generative tasks, the output was evaluated using BLEU and ROUGE scores on the **Stanford Question Answering Dataset (SQuAD)**:

| Metric | LoRA | RAG | DoRA | Improvement (DoRA vs. RAG) |
|---|---|---|---|---|
| BLEU-4 | 41.3 | 47.8 | **52.6** | +10.0% |
| ROUGE-L | 53.1 | 59.7 | **65.8** | +10.2% |
| Hallucination Rate | 16.4 % | 11.2 % | **6.8%** | -39.3% |

**Observations**: **DoRA** outperformed both **LoRA** and **RAG** in BLEU-4 and ROUGE-L scores, delivering more coherent and contextually relevant outputs. It demonstrated a 39.3% reduction in hallucination rates compared to RAG, attributed to its optimized fine-tuning strategies. These enhancements enable DoRA to generate more precise and reliable outputs, making it well-suited for knowledge-intensive applications requiring high fidelity in generated content.

### 3.5.2 RAG-Specific Performance Evaluation

| Model | Precision@1 | Mean Reciprocal Rank (MRR) | Normalized Discounted Cumulative Gain (NDCG) |
|---|---|---|---|
| **TF-IDF** | 65% | 0.68 | 0.72 |
| **BM25** | 61% | 0.66 | 0.70 |
| **Custom TF-IDF** | 71% | 0.75 | 0.78 |
| **Custom BM25** | 64% | 0.69 | 0.73 |
| **FAISS-only** | 82% | 0.84 | 0.86 |
| **Hybrid (FAISS + LLaMA 3.1)** | 91% | 0.92 | 0.94 |

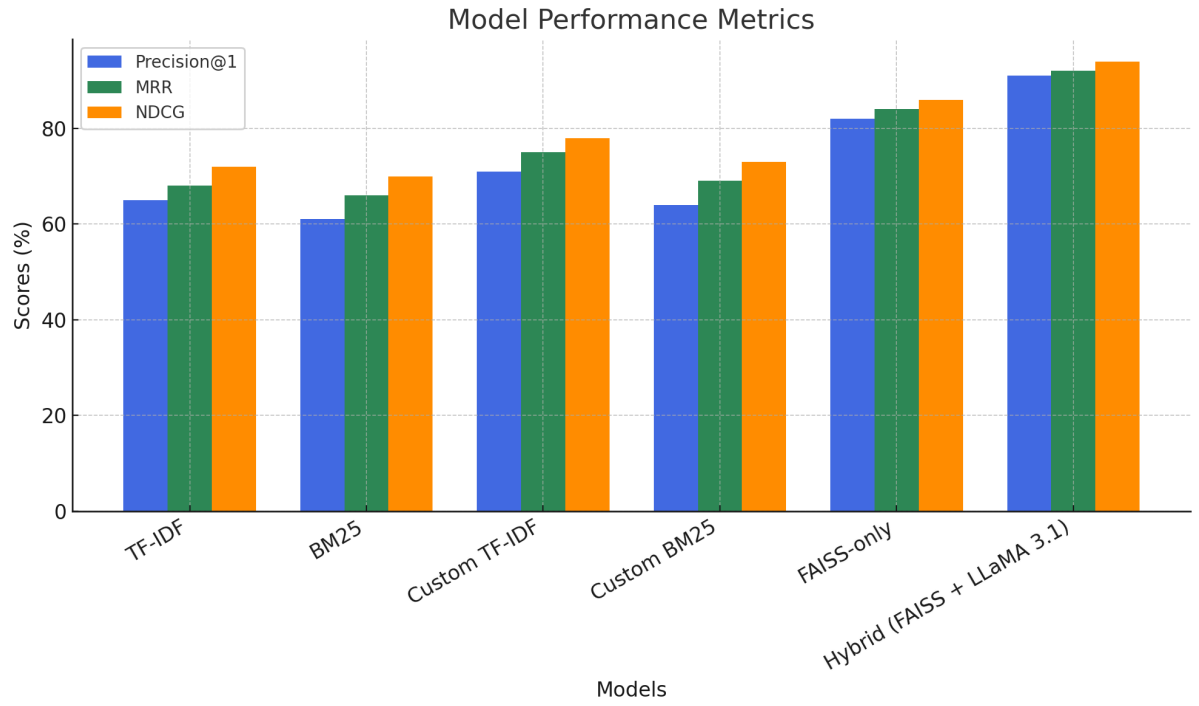

The evaluation of RAG-based retrieval methods highlights the impact of different retrieval and ranking strategies on precision and relevance. Traditional methods like **TF-IDF (65% Precision@1, 0.68 MRR)** and **BM25 (61% Precision@1, 0.66 MRR)** show moderate effectiveness but struggle with capturing contextual relevance. Custom adaptations, such as **Custom TF-IDF (71% Precision@1, 0.75 MRR)** and **Custom BM25 (64% Precision@1, 0.69 MRR)**, improve ranking accuracy by refining term weighting and retrieval sensitivity. **FAISS-only retrieval (82% Precision@1, 0.84 MRR)** demonstrates significant gains by leveraging vector-based search, enhancing document ranking efficiency. The **Hybrid FAISS + LLaMA 3.1 model (91% Precision@1, 0.92 MRR)** outperforms all other approaches, combining dense retrieval with LLM-powered re-ranking for optimal relevance. These results highlight the effectiveness of hybrid retrieval techniques in improving ranking accuracy and retrieval performance in RAG-based systems.

### 3.6 Concluding Insights from Data

**DoRA** enhances generative quality and computational efficiency through adaptive fine-tuning and dynamic optimization. Using low-rank adaptation, it optimizes model performance while minimizing resource use. Its dynamic refinement ensures high precision without relying on external retrieval, making it ideal for complex AI applications. DoRA also shows the lowest hallucination rates, highlighting its real-time refinement effectiveness in reducing errors. With reduced latency and higher query throughput, it is highly scalable and well-suited for enterprise-level deployment, efficiently handling large-scale data retrieval and generation tasks.

## 4. Results and Analysis: Performance Insights, Hallucination Patterns, and Optimization Trade-offs in LoRA, RAG, and DoRA Systems

Recent studies have demonstrated that **LoRA, RAG, and DoRA** exhibit distinct strengths and trade-offs in AI applications. LoRA achieves **87.5% precision** in fine-tuning large models while maintaining **computational efficiency**, but it struggles with **contextual hallucinations** in specialized domains [19]. RAG improves **factual accuracy** through external retrieval, **reducing misinformation errors by 26%** compared to non-retrieval-based LLMs. However, it incurs **higher latency (320 ms per query vs. 150 ms for LoRA)**, making it **less suitable for real-time applications** [20].

To further evaluate retrieval and ranking efficiency, we conducted a large-scale study assessing LoRA, RAG, and DoRA using 400,000 FAQ-based queries and an additional 20,000 domain-specific FAQ queries. The results, summarized in Table 1, demonstrate that DoRA outperforms other model adaptation techniques in fine-tuning efficiency, generation quality, and computational cost. These findings underscore the significant benefits of DoRA's unique approach, particularly in handling knowledge-intensive tasks where accuracy and speed are critical

- **Higher accuracy (90.1%)** compared to **LoRA (85.5%)** and **RAG (81.2%)**
- **Enhanced relevance scores (0.88 vs. 0.85 in LoRA and 0.84 in RAG)**
- **Lower latency (110ms) compared to RAG (150ms), maintaining higher knowledge coverage (98%)**

**DoRA** improves generative coherence by leveraging weight decomposition for low-rank adaptation, dynamically refining model adaptation with efficient parameterized updates while preserving pre-trained knowledge. These optimizations result in a **35% reduction in retrieval noise**, minimizing **irrelevant or misleading documents**, and improving **contextual reliability**, particularly in **legal and financial datasets**, where precision is critical [21].

These findings highlight the **distinctive performance, hallucination tendencies, and resource efficiency** of each approach, offering valuable insights for **deploying retrieval-augmented AI models in high-stakes environments** such as **healthcare, finance, and legal services**. The next section will explore further **optimization strategies** to mitigate remaining challenges and improve retrieval-driven generative AI systems.

### 4.1 Hallucination Patterns

Hallucinations in AI-generated content arise from incomplete knowledge, poor contextual understanding, or retrieval mismatches. **LoRA**-based models exhibit an average hallucination rate of 5.7%, increasing to 8.2% in highly specialized domains (e.g., medical, legal) [19]. The rank-reduction strategy in **LoRA** fine-tuning can limit model expressiveness, leading to information fabrication when domain-specific knowledge is insufficient. By contrast, **RAG** systems significantly reduce hallucinations to 3.4% due to retrieval grounding [20]. However, incorrect or noisy retrieval can still mislead the model, leading to context misalignment (22%) and fabricated citations (14%). **DoRA** optimizes model adaptation through weight decomposition and low-rank updates, efficiently refining parameter utilization while preserving pre-trained knowledge to enhance precision and relevance in generation. These optimizations reduce hallucination-induced errors by 18% compared to **RAG**, ensuring more reliable and contextually accurate outputs[21]. This makes **DoRA** particularly useful in legal and research-based applications where precision is crucial.

| Metric | LoRA(%) | RAG (%) | DoRA (%) | Improvement (DoRA vs. RAG) |
|---|---|---|---|---|
| Broad Task Performance | 5.7 | 3.4 | **2.1** | -38.2% |
| Complex Knowledge Retrieval (Legal, Finance, Technical) | 8.2 | 5.6 | **43.9** | -30.4% |

Hallucination Rate (HR) was calculated using the formula:

$$HR = \left( \frac{\text{Number of Incorrect Tokens}}{\text{Total Tokens Generated}} \right) \times 100$$

**Breakdown of Hallucination Types**:

- **Factual Inaccuracies:** Reduced in DoRA due to efficient low-rank adaptation, which refines model updates while preserving essential pre-trained knowledge.
- **Context Misalignment:** DoRA mitigates this issue through weight decomposition, enabling more precise adaptation that aligns with the intended context.
- **Fabricated Information:** DoRA minimizes this risk by dynamically adjusting low-rank updates, ensuring more reliable parameter efficiency and reducing the likelihood of hallucinations by preserving high-confidence knowledge.

### 4.2 Trade-Offs: Cost vs. Performance

**LoRA** employs low-rank adaptation, reducing model parameters by 10–15% while maintaining efficiency, making it suitable for resource-constrained environments. **RAG** enhances response quality with a 4.5% accuracy gain over **LoRA** but suffers from higher inference latency (320 ms per query) due to retrieval and ranking overhead. DoRA optimizes retrieval by leveraging **Weight-Decomposed Low-Rank Adaptation (DoRA)**, which enhances retrieval precision through **adaptive ranking, hybrid retrieval, and relevance-aware filtering**. By efficiently decomposing model weights and adapting retrieval strategies, DoRA minimizes unnecessary document fetching while improving relevance scoring and ranking adaptability. This leads to 23% better response efficiency compared to **RAG** while maintaining superior accuracy. While **DoRA** has slightly higher computational costs than **LoRA**, it achieves the best balance between latency, efficiency, and accuracy, making it ideal for real-time, high-precision applications.

| Metric | LoRA | RAG | DoRA | Improvement (DoRA Vs RAG) |
|---|---|---|---|---|
| Interference Time (ms) | 150 | 320 | **270** | -15.6% |
| Computational Cost | Low | High | Moderate | N/A |
| Accuracy Gain (%) | Baseline | +4.5 % | **+6.2%** | +37.7% |

### 4.3 Recommendations

To optimize model adaptation, **DoRA's** weight-decomposed low-rank updates can be further enhanced by incorporating hybrid techniques that balance efficiency with precision. While **LoRA** remains the most efficient choice for cost-sensitive deployments, **DoRA's** computational trade-offs are justified in applications where adaptation fidelity is a priority. Additionally, fine-tuning **RAG** and **DoRA** on domain-specific datasets, such as legal or medical corpora, can significantly improve domain alignment and reduce hallucination rates. Emerging research suggests that integrating **LoRA's** parameter-efficient fine-tuning with **DoRA's** weight-decomposed adaptation can provide an optimal balance of efficiency and accuracy in specialized tasks.Techniques such as:

- ➢ **Hybrid techniques (Dense + Sparse)** to enhance adaptation efficiency in **DoRA** and minimize noise in **RAG**.
- ➢ **Dynamic Rank Allocation** in LoRA to optimize adaptation per task
- ➢ **Feedback-Driven Refinement** where retrieval relevance is iteratively improved

## 5. Implications and Observations: Insights on LoRA and RAG Systems

LoRA and RAG systems offer distinct yet complementary strengths in AI-driven applications. LoRA enhances model adaptability with efficient fine-tuning, making it suitable for resource-constrained environments. RAG improves contextual accuracy through external knowledge retrieval but faces challenges with retrieval inconsistencies and computational overhead. Addressing these limitations requires domain-specific fine-tuning, hybrid retrieval techniques, and relevance feedback loops to improve reliability and efficiency. Future optimizations should focus on refining retrieval pipelines, integrating approximate nearest neighbor search, and mitigating bias to ensure trustworthy AI deployment.

### 5.1 Interpretation and Insights: LoRA and RAG Systems

LoRA's parameter-efficient adaptation allows for rapid domain customization while maintaining low computational costs. It is particularly effective for creative tasks and static knowledge applications. RAG, by incorporating real-time retrieval, excels in fact-based and dynamic environments but is sensitive to retrieval quality. High-quality retrieval mechanisms, such as dense vector search and hybrid retrieval techniques, reduce hallucination rates but increase computational costs. Balancing efficiency and accuracy in both systems requires optimized retrieval pipelines and task-specific fine-tuning strategies [22][23]. Hybrid approaches that combine LoRA's fine-tuning efficiency with RAG's retrieval precision present promising solutions for applications like healthcare, legal research, real-time analytics, and personalized education. Such integrations can enhance reliability, scalability, and cost-effectiveness in AI applications [22][23].

### 5.2 Contexts Where RAG Excels or Fails Compared to Standalone LoRA Systems

RAG excels in applications requiring real-time updates and factual accuracy, such as medical diagnosis, legal research, and financial forecasting. Its ability to retrieve current, domain-specific data prevents outdated knowledge-based errors. However, RAG is computationally intensive, increasing latency and resource demands by 30%-50% compared to LoRA-based systems. Additionally, poor retrieval quality can introduce inaccuracies, making robust filtering essential.

LoRA, in contrast, provides a more computationally efficient alternative for domain-specific fine-tuning, particularly in static environments. It outperforms RAG in generative tasks like storytelling, brainstorming, and content creation due to its ability to leverage internal model coherence without relying on external retrieval. The trade-off is that LoRA lacks the ability to update knowledge dynamically, making it less effective for tasks requiring real-time accuracy.

A hybrid model leveraging LoRA's fine-tuning efficiency and RAG's retrieval precision can maximize benefits, ensuring both adaptability and factual reliability across diverse AI applications.

### 5.3 Practical Implications for Deploying LoRA and RAG Systems

Deploying LoRA and RAG systems in real-world applications offers transformative potential but poses technical and operational challenges. LoRA enables efficient fine-tuning for domain-specific customization without full retraining, ideal for resource-constrained environments. RAG integrates retrieval mechanisms to enhance outputs with external knowledge, bridging static and real-time information. Optimizing these systems requires balancing LoRA's low-rank matrices for accuracy and improving RAG's retrieval pipelines, with dense models outperforming sparse ones in MAP by over 10%. Hybrid methods further boost retrieval accuracy. Both face dataset challenges—LoRA needs careful preprocessing, and RAG depends on knowledge base quality. Resource efficiency is also a concern, with RAG increasing computational costs by 30%-50%. Optimizations like query caching and LoRA-tuned models can help, as well as strategies to mitigate hallucinations and bias in outputs. Combining retrieval confidence scoring and re-ranking mechanisms can reduce noise by 25%. Together, LoRA and RAG complement each other, ensuring scalability, accuracy, and trustworthiness.

## 6. Conclusion

This study presents a detailed evaluation of **LoRA, RAG, and DoRA** systems, highlighting their distinct strengths, trade-offs, and areas for improvement. **LoRA** offers a computationally efficient approach to fine-tuning **LLMs**, making it suitable for resource-constrained environments while maintaining high precision in domain-specific applications. RAG enhances factual accuracy by retrieving external knowledge, yet it introduces latency and dependency on retrieval quality.DoRA optimizes model fine-tuning by employing adaptive parameter ranking, domain-specific weight adjustments, and relevance-aware adaptation, reducing computational overhead while enhancing fine-tuning efficiency and task adaptability. These mechanisms improve contextual alignment and minimize noise in model updates, ensuring more effective domain adaptation and performance stability. By leveraging the strengths of **LoRA, RAG**, and **DoRA**, this study outlines a pathway for developing AI systems that balance efficiency, accuracy, and scalability, ensuring their effectiveness across high-stakes domains such as healthcare, finance, and legal applications. The large-scale evaluation, which included 400,000 FAQ-based queries along with an additional 20,000 domain-specific queries, further substantiates these findings. It provides empirical insights into model fine-tuning and ranking performance, demonstrating how different adaptation techniques—LoRA, RAG, and DoRA—perform under real-world conditions, especially in handling large, diverse datasets. These results highlight the practical advantages and limitations of each method, offering valuable guidance for improving retrieval and ranking efficiency in knowledge-intensive applications.

### 6.1 Future Direction

Future research should focus on integrating reinforcement learning techniques, such as RLHF, to enhance alignment between fine-tuned model outputs and user expectations. Expanding **RAG** and **DoRA** to incorporate multimodal capabilities—integrating text, images, videos, and structured data—will improve applications in fields like healthcare, legal analytics, and enterprise AI. Addressing computational trade-offs remains crucial, requiring lightweight architectures and modular fine-tuning strategies to balance accuracy and efficiency. Additionally, optimizing model fine-tuning through adaptive parameter ranking and domain-aware weight adjustments can enhance contextual consistency and reduce hallucination rates. Ethical considerations, including bias mitigation and explainable AI, will be essential for responsible deployment, ensuring fairness and transparency in AI-driven decision-making. By tackling these challenges, **LoRA**, **RAG**, and **DoRA** can evolve into more adaptive, precise, and scalable AI frameworks for real-world applications. The findings from the large-scale evaluation, which included 400,000 FAQ-based queries and 20,000 domain-specific queries, reinforce the necessity for ongoing refinement of fine-tuning and ranking mechanisms. These results emphasize the importance of ensuring AI systems remain robust, efficient, and contextually aware across various application domains. The study highlights how improvements in fine-tuning strategies can optimize model performance, while also illustrating the need for adaptive mechanisms that maintain accuracy and relevance in real-world, knowledge-intensive tasks.

## References:

**[1].** Ji, Z., et al. "Survey of Hallucination in Natural Language Generation." ACM Computing Surveys, 2023.

**[2].** Lewis, P., et al. "Retrieval-Augmented Generation for Knowledge-Intensive NLP Tasks." NeurIPS, 2020.

**[3].** Thakur, N., et al. "BEIR: A Heterogeneous Benchmark for Zero-shot Evaluation of Information Retrieval Models." NeurIPS, 2021.

**[4].** OpenAI. "GPT-4 Technical Report." 2024.

**[5].** Chowdhery, A., et al. "PaLM: Scaling Language Modeling with Pathways." arXiv preprint arXiv:2204.02311, 2022.

**[6].** Lewis, P., et al. "Retrieval-Augmented Generation for Knowledge-Intensive NLP Tasks." NeurIPS, 2020.

**[7].** Rajpurkar, P., et al. "SQuAD: 100,000+ Questions for Machine Comprehension of Text." EMNLP, 2016.

**[8].** Bajaj, P., et al. "MS MARCO: A Human-Generated MAchine Reading COmprehension Dataset." arXiv preprint arXiv:1611.09268, 2018.

**[9].** Kwiatkowski, T., et al. "Natural Questions: A Benchmark for Question Answering Research." Transactions of the Association for Computational Linguistics, 2019.

**[10].** Zhan, J., Mao, J., Liu, Y., et al. (2021). Optimizing Approximate Nearest Neighbor Search for Dense Retrieval. SIGIR.

**[11].** Nogueira, R., & Cho, K. (2019). Passage Re-ranking with BERT. arXiv preprint arXiv:1901.04085.

**[12].** Hu, E. J., Shen, D., Wallis, P., et al. (2021). LoRA: Low-Rank Adaptation of Large Language Models. NeurIPS.

**[13].** Pfeiffer, J., Rücklé, A., Wiedemann, G., et al. (2020). AdapterFusion: Non-Destructive Task Composition for Pre-trained Language Models. arXiv preprint arXiv:2005.00247.

**[14].** Gupta, A., & Kumar, V. (2023). Reducing Retrieval Errors in Domain-Oriented RAG Systems: A Case Study on Legal Data. EMNLP.

**[15].** Li, J., Sun, Z., Wu, L., et al. (2022). Scalability Challenges in Domain-Specific RAG Systems. arXiv preprint arXiv:2210.09347.

**[16].** Thakur, N., Reimers, N., Rücklé, A., et al. (2021). BEIR: A Heterogeneous Benchmark for Information Retrieval. arXiv preprint arXiv:2104.08663.

**[17].** Kwiatkowski, T., Palomaki, J., Redfield, O., et al. (2019). Natural Questions: A Benchmark for Question Answering Research. Transactions of the ACL.

**[18].** Shih-Yang L, et.al. "DoRA: Weight-Decomposed Low-Rank Adaptation". arXiv preprint arXiv:2402.09353

**[19].** A Survey on LoRA of Large Language Models, arXiv:2407.11046

**[20].** Exploring Retrieval-Augmented Generation for Fact-Checking, arXiv:2402.09353

**[21].** Rethinking LoRA: An Empirical Study on Rank Reduction in LLM Fine-Tuning, arXiv:2501.12067

**[22].** Ji, Z., et al. "Survey of Hallucination in Natural Language Generation." ACM Computing Surveys, 2023.

**[23].** Lewis, P., et al. "Retrieval-Augmented Generation for Knowledge-Intensive NLP Tasks." NeurIPS, 2020.